\documentclass{article}
\usepackage[latin9]{inputenc}
\usepackage{booktabs}
\usepackage{url}
\usepackage{graphicx}
\usepackage[unicode=true,
 bookmarks=false,
 breaklinks=false,pdfborder={0 0 1},backref=section,colorlinks=false]
 {hyperref}

\makeatletter

\providecommand{\tabularnewline}{\\}

\usepackage{times}\usepackage{url}

\title{Challenging Images For Minds and Machines}

\author{Amir Rosenfeld, John K. Tsotsos  \\
Department of Electrical Engineering and Computer Science\\ York University, Toronto, ON, Canada\\
\texttt{amir@eecs.yorku.ca,tsotsos@cse.yorku.ca} \\
}

\makeatother

\begin{document}
\maketitle
\begin{abstract}
There is no denying the tremendous leap in the performance of machine
learning methods in the past half-decade. Some might even say that
specific sub-fields in pattern recognition, such as machine-vision,
are as good as solved, reaching human and super-human levels. Arguably,
lack of training data and computation power are all that stand between
us and solving the remaining ones. In this position paper we underline
cases in vision which are challenging to machines and even to human
observers. This is to show limitations of contemporary models that
are hard to ameliorate by following the current trend to increase
training data, network capacity or computational power. Moreover,
we claim that attempting to do so is in principle a suboptimal approach.
We provide a taster of such examples in hope to encourage and challenge
the machine learning community to develop new directions to solve
the said difficulties. 
\end{abstract}

\section{Introduction\label{intro}}

\begin{figure}[h]
\centering{}\includegraphics[width=0.65\textwidth]{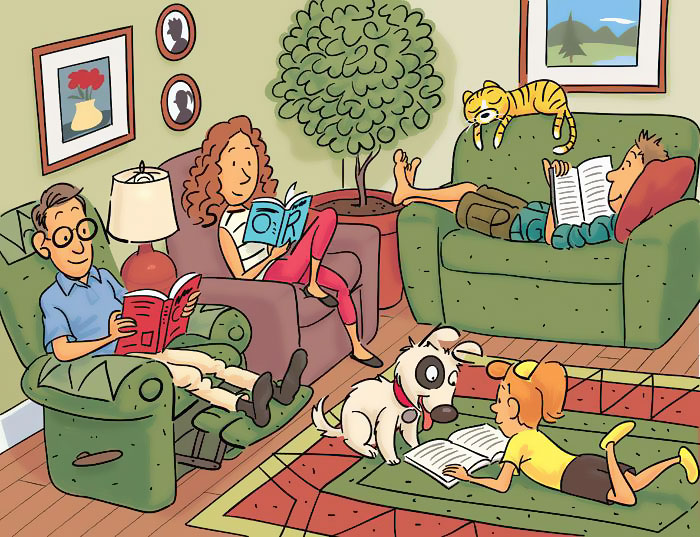}\caption{\label{fig: hidden-words}A children's puzzle where the goal is to
find six hidden words: Book, words, story, pages, read, novel. For
a machine this is far from child's play. Could this be solved by providing
a million similar examples to a deep-learning system? Does a human
need such training?}
\end{figure}

Once only known to a few outside of academia, machine-learning has
become ubiquitous in both popular media and in the industry. Superhuman
capabilities are now being gradually recorded in various fields: in
the game of GO, (\cite{silver2016mastering,silver2017mastering}),
in face verification (\cite{lu2015surpassing,qi2018face}), image
categorization (\cite{he2015delving}) and even in logical reasoning
in simple scenes (\cite{santoro2017simple,perez2017learning,perez2017film}).

Most current leading methods involve some variant of deep learning.
Consequentially, they require large amounts of hand-labeled data (with
the exception of \cite{silver2017mastering} - which used self-play
to gain experience). This has elicited a data-hungry era, with increasingly
large-scale datasets painstakingly labeled for object classification/detection/segmentation,
image annotation, visual question-answering, and pose estimation (\cite{russakovsky2015imagenet,lin2014microsoft,krishna2017visual,antol2015vqa,guler2018densepose})
to name a few. This is accompanied by a growing demand for computational
power.

We bring forward challenges in vision which do not seem to be solved
by current methods - and more importantly - by current popular methodologies,
meaning that neither additional data, nor added computational power
will be the drivers of the solution. 

\subsection*{Related Work\label{sec:Related-Work}}

\textbf{Imbalanced or Small Data: }datasets tend to be naturally imbalanced,
and there is a long history of suggested remedies (\cite{lim2011transfer,zhu2014capturing,wang2017learning}).
Handling lack of training data has also been treated by attempting
to use web-scale data of lesser quality than hand-annotated dataset
\cite{sun2017revisiting}, simulating data {[}cite data for cars,
text recognition in the wild, captcha{]}. \textbf{Transfer Learning:
}reusing features of networks trained on large is a useful starting
point (cf \cite{sharif2014cnn}) \textbf{One-Shot-Learning}: attempting
to reduce the number of required training example, in extreme cases
to one or even zero examples (\cite{snell2017prototypical});\textbf{
Deep-Learning Failures}: recently, some simple cases where deep learning
fails to work as one would possibly expect were introduced, along
with theoretical justifications (\cite{shalev2017failures}). 

\section{Challenging Cases}

We present two examples and then discuss them. They have a few common
characteristics: humans are able to solve them on the first ``encounter''
- despite not having seen any such images before. Incidentally - but
not critically - the two examples are from the domain of visual text
recognition. Moreover, though humans know how to recognize text as
seen in regular textbooks, street-signs, etc, the text in these images
is either hidden, rendered, or distorted in an uncharacteristic manner. 

\textbf{Children's games}: the first case is well exemplified by a
child's game, hidden word puzzles. The goal is to find hidden words
in an image. Fig. \ref{fig: hidden-words} shows an arbitrarily selected
example. For a human observer this is a solvable puzzle, though it
may take a few minutes to complete. We applied two state-of-the-art
methods for text recognition in the wild with available code (\cite{shi2017end})
or an on line-demo (\cite{zhou2017east}\footnote{\url{http://east.zxytim.com}})
on the image in Fig. \ref{fig: hidden-words}. As this did not work
immediately, we focused on the word ``NOVEL'' (the ``N'' is below
the forearm of the left person, ending with an ``L'' below his foot),
by cropping it an rotating so the text is level, cropping more tightly
, and even cropping only the letter ``L''. See Table \ref{tab:crops}
for the corresponding sub-images (including the entire image at the
top row) and the results output by the two methods. 

This is by no means a systematic test and some may even claim that
it isn't fair - and they would be right: these systems were not trained
on such images; \cite{shi2017end} was only trained on a photo-realistic
dataset of 8 million synthetic training images, and \cite{zhou2017east}
was only trained on tens of thousands of images from coco-text (\cite{veit2016coco}),
or used powerful pre-trained networks where training data was less
available. 
\begin{table}
\begin{centering}
\begin{tabular}{ccccc}
Sub Image & \includegraphics[height=0.07\textheight]{figures/find-words-challenge-18__700} & \includegraphics[height=0.07\textheight]{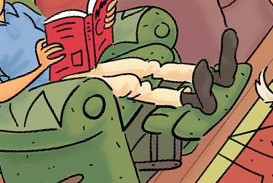} & \includegraphics[height=0.07\textheight]{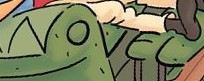} & \includegraphics[height=0.07\textheight]{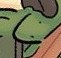}\tabularnewline
\midrule 
\cite{shi2017end} & ``sned'' & ``vvoz'' & ``novees'' & ``teg''\tabularnewline
\midrule 
\cite{zhou2017east} & ``score'' & $\emptyset$ & $\emptyset$ & $\emptyset$\tabularnewline
\bottomrule
\end{tabular}
\par\end{centering}
\caption{\label{tab:crops}Text detected by two state-of-the-art scene-text
recognition methods applied to sub-images of a children's puzzle.
$\emptyset$ means no text was detected by the method (images scaled
to fit figure). }
\end{table}

\begin{figure}
\begin{centering}
\includegraphics[bb=0bp 15bp 1592bp 76bp,clip,width=0.75\textwidth]{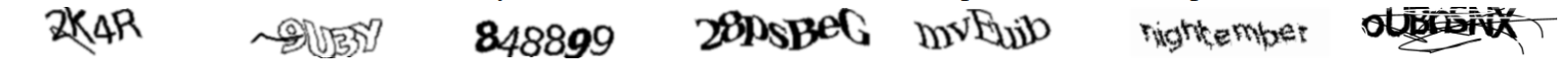}
\par\end{centering}
\caption{Variants of textual CAPTCHA. Captchas are becoming increasingly difficult
(reproduced from \cite{le2017using}) }

\end{figure}
\textbf{CAPTCHA}: a well-known mechanism to thwart automated misuse
of websites by distinguishing between humans and machines (\cite{von2003captcha}).
Textual captchas involve presenting an image of text which has to
be read and written by the user. We focus on this type of captcha,
though others exist (\cite{singh2014survey}). The introduction of
captchas immediately triggered the invention of new automatic ways
to break them (\cite{mori2003recognizing}), which eventually sparked
an ``arms race'' between increasingly complex captchas and correspondingly
powerful automated methods (\cite{chen2017survey}). This caused a
state where on one-hand the best leading textual captcha-solution
methods involve training DNN's over data with similar distortion characteristics
as the desired types of captcha - though still these systems have
limited success rates (at times less than 50\%) - and on the other
hand the level of distortion has become such that humans have a hard-time
solving some of them. 

\section{Machines vs Humans as Supervised Learners}

One can rule out the suggested examples by saying that they are simply
out-of-sample datapoints on behalf of a statistical learner's perspective.
Yet it seems that with whatever supervision human-beings receive -
they are usually able to solve them despite not being especially exposed
to this kind of stimulus. Moreover, precisely these kinds of images
are used routinely in human IQ testing, so they are a universally
accepted indicator for human performance. If these examples may seem
esoteric, we can revert to more common cases: as a child, how often
is one exposed to bounding boxes of objects? How often to delineations
of objects with precise segmentation masks? How often to pose-configurations,
facial and bodily key-points, and dense-meshes of 3D objects overlayed
on their field of view (\cite{guler2018densepose})? More critically,
for how many different object types does this happen (if any), for
how many different instances, with what level of precision of annotation,
and in how many modalities?

The granularity of visual supervision given to machines seems to be
much finer than that given to humans. As for the amount of directly
supervised data, it does not seem to really be the main limiting factor;
as already noted several times, performance either saturates with
training data (\cite{zhu2012we,zhu2016we}) or at best grows logarithmically
(\cite{sun2017revisiting,hestness2017deep}, increasing mAP from 53\%
to 58\% when growing from 10M to 300M examples) making the solution
of more data for better performance simply impractical - even for
those with the most resources. And this is for ``common'' problems,
such as object detection. 

Humans who only ever read street-signs and textbooks are able to solve
captchas of various kinds without any special training on their first
encounter with them. The same is true for the ``picture puzzles''
mentioned above, as it is for other cases not mentioned here. We do
not claim that humans are not subject to supervised learning in their
early life, and in later stages. On the contrary, supervisory signals
arise from multiple sources: caretakers who provide supervisory signals
by teaching, ``internal supervision'' provided by innate biases
(\cite{ullman2012simple}) and finally rewards stemming from results
of behaviour, such as suffering pain from hitting an object. But any
such supervision is interspersed within a vast, continuous stream
of unsupervised data, most of which does not have an easily measurable
supervisory affect on the observer. 

There is something fundamentally different about the way humans construct
or use internal representations, enabling them to reason about and
solve new pattern-recognition tasks. We hypothesize that these are
approached by generating procedures of a compositional nature when
presented with a novel - or known - task (as suggested by the Visual
Routines of \cite{ullman1984visual} or the Cognitive Programs of
\cite{tsotsos2014cognitive}. We intend to maintain a collection of
examples beyond the ones suggested above, to encourage the community
to attempt to solve them, not by learning from vast amounts of similar
examples, but by learning from related, simpler subtasks and learning
to reason and solve them by composing the appropriate solutions. 

\bibliographystyle{IEEEtran}
\bibliography{iclr2018_workshop_2}

\end{document}